\documentclass[preprint]{article}
\usepackage{neurips_2025}

\usepackage[utf8]{inputenc} 
\usepackage[T1]{fontenc}    
\usepackage{graphicx}       
\usepackage{hyperref}       
\usepackage{url}            
\usepackage{booktabs}       
\usepackage{amsfonts}       
\usepackage{nicefrac}       
\usepackage{microtype}      
\usepackage{lipsum}		
\usepackage{graphicx}
\usepackage{natbib}
\usepackage{doi}
\usepackage{enumitem}
\usepackage{amsmath}
\usepackage{cases}
\usepackage{subcaption} 
\usepackage{booktabs} 
\usepackage{multirow}   
\usepackage{array}      
\usepackage{ifthen}
\usepackage[ruled,vlined]{algorithm2e}
\usepackage{algpseudocode}
\usepackage[table]{xcolor}
\usepackage{multicol}
\usepackage{geometry}
\usepackage{xfp}

\usepackage{cleveref}
\crefname{section}{§}{§§}
\Crefname{section}{§}{§§}

\usepackage{neurips_2025}
\usepackage[utf8]{inputenc}
\usepackage[T1]{fontenc}
\usepackage{hyperref}
\usepackage{url}
\usepackage{booktabs}
\usepackage{amsfonts}
\usepackage{nicefrac}
\usepackage{microtype}
\usepackage{algorithmicx}
\usepackage{algpseudocode}
\usepackage{graphicx}
\usepackage{wrapfig}
\usepackage{caption}
\usepackage{times}
\usepackage{latexsym}
\usepackage[T1]{fontenc}
\usepackage[utf8]{inputenc}
\usepackage{inconsolata}
\usepackage{amsmath}
\usepackage{amssymb}
\usepackage{enumerate}
\usepackage{balance}
\usepackage{blindtext}
\usepackage{colortbl}
\usepackage{float}
\usepackage{makecell}
\usepackage{multicol}
\usepackage{multirow}
\usepackage{tabularx}
\usepackage{marvosym}
\usepackage{listings}
\usepackage{tcolorbox}
\usepackage{stfloats}
\usepackage{subcaption}
\usepackage{ulem}
\usepackage{bm}
\usepackage{longtable}
\usepackage{array}
\usepackage{colortbl}
\usepackage{siunitx}

\usepackage[utf8]{inputenc} 
\usepackage[T1]{fontenc}    
\usepackage{hyperref}       
\usepackage{url}            
\usepackage{booktabs}       
\usepackage{amsfonts}       
\usepackage{nicefrac}       
\usepackage{microtype}

\newcommand\footnoteONLYtext[1]{
    \let \mybackup \thefootnote
    \let \thefootnote \relax
    \footnotetext{#1}
    \let \thefootnote \mybackup
    \let \mybackup \imareallyundefinedcommand}

\title{Cross-Task Experiential Learning on LLM-based Multi-Agent Collaboration}

\author{
\textbf{Yilong Li}{\footnotesize $^\triangle$\textsuperscript{\textdagger}} \quad
\textbf{Chen Qian}{\footnotesize $^\clubsuit$\textsuperscript{\textdagger}} \quad
\textbf{Yu Xia}{\footnotesize $^\bigstar$} \quad
\textbf{Ruijie Shi}{\footnotesize $^\bigstar$} \quad
\textbf{Yufan Dang}{\footnotesize $^\bigstar$} \quad \\
\textbf{Zihao Xie}{\footnotesize $^\bigstar$} \quad 
\textbf{Ziming You}{\footnotesize $^\triangle$} \quad 
\textbf{Weize Chen}{\footnotesize $^\bigstar$} \quad 
\textbf{Cheng Yang}{\footnotesize $^\spadesuit$} \quad 
\textbf{Weichuan Liu}{\footnotesize $^\diamondsuit$} \quad
\textbf{Ye Tian}{\footnotesize $^\heartsuit$} \quad \\
\textbf{Xuantang Xiong}{\footnotesize $^\heartsuit$} \quad 
\textbf{Lei Han}{\footnotesize $^\heartsuit$} \quad 
\textbf{Zhiyuan Liu}{\footnotesize $^\bigstar$\textsuperscript{\Letter}} \quad
\textbf{Maosong Sun}{\footnotesize $^\bigstar$\textsuperscript{\Letter}} \quad \\
{\footnotesize $^\triangle$}Peking University \quad
{\footnotesize $^\clubsuit$}Shanghai Jiao Tong University \quad \\
{\footnotesize $^\bigstar$}Tsinghua University \quad
{\footnotesize $^\spadesuit$}Beijing University of Posts and Telecommunications \quad \\
{\footnotesize $^\diamondsuit$}Siemens \quad 
{\footnotesize $^\heartsuit$}Tencent Robotics X \quad \\
\href{liyilong@stu.pku.edu.cn}{\texttt{liyilong@stu.pku.edu.cn}} \quad 
\href{qianc@sjtu.edu.cn}{\texttt{qianc@sjtu.edu.cn}} \\
\href{liuzy@tsinghua.edu.cn}{\texttt{liuzy@tsinghua.edu.cn}} \quad 
\href{sms@tsinghua.edu.cn}{\texttt{sms@tsinghua.edu.cn}}
}

\begin{document}

\maketitle

\footnoteONLYtext{$^\dagger$Equal Contribution.}
\footnoteONLYtext{$^{\text{\Letter}}$Corresponding Author.}

\begin{abstract}

  Large Language Model–based multi-agent systems (MAS) have shown remarkable progress in solving complex tasks through collaborative reasoning and inter-agent critique. However, existing approaches typically treat each task in isolation, resulting in redundant computations and limited generalization across structurally similar tasks. To address this, we introduce multi-agent cross-task experiential learning (MAEL), a novel framework that endows LLM-driven agents with explicit cross-task learning and experience accumulation. 
  We model the task‐solving workflow on a graph‐structured multi-agent collaboration network, where agents propagate information and coordinate via explicit connectivity.
  During the experiential learning phase, we quantify the quality for each step in the task-solving workflow and store the resulting rewards along with the corresponding inputs and outputs into each agent's individual experience pool.
  During inference, agents retrieve high-reward, task-relevant experiences as few-shot examples to enhance the effectiveness of each reasoning step, thereby enabling more accurate and efficient multi-agent collaboration.
  Experimental results on diverse datasets demonstrate that MAEL empowers agents to learn from prior task experiences effectively—achieving faster convergence and producing higher‑quality solutions on current tasks.
\end{abstract}

\section{Introduction}
Large Language Models (LLMs) have pushed the frontier of natural language processing by demonstrating strong zero-/few-shot generalization, in-context reasoning, and tool use capabilities \citep{vaswani2017attention,radford2019language,qin2023toolllm,schick2023toolformer}. 
To harness these capabilities for complex tasks, recent research has explored Multi-Agent Systems (MAS) in which multiple LLM-based agents collaborate—each specializing in reasoning, planning, or execution—to decompose and solve problems more effectively than a single model could achieve. 
Early MAS frameworks such as AgentVerse \citep{chen2023agentverse} and CAMEL \citep{li2023camel} pioneered modular task decomposition and parallel collaboration paradigms, demonstrating how groups of specialized LLM agents can jointly tackle complex problems. Subsequent systems like AutoGen \citep{wu2023autogen} and MetaGPT \citep{hong2023metagpt} introduced flexible interaction protocols and SOP‑guided role adaptation, enabling agents to fluidly switch among planning, coding, and evaluation modes. More recent contributions have integrated inter‑agent critique loops—mechanisms by which agents systematically review, challenge, and improve one another’s intermediate outputs—to enhance error detection, reinforce consistency, and progressively elevate solution quality \citep{park2023generative, liu2024dynamic}.

However, existing MAS frameworks typically operate in a myopic fashion: during inference, agents collaborate to solve each task from scratch, without leveraging knowledge or experience gained from previously solved tasks. This limitation often results in repetitive trial-and-error when encountering new tasks that share structural similarities with past ones, leading to suboptimal sample efficiency and longer convergence time. While some self-evolving methods have explored dynamic adaptation of agent collaboration topologies at test-time \citep{hu2025selfevolving,zhang2025evoflow}, they do not explicitly accumulate or transfer inter-task experience through a dedicated learning phase.

In this paper, we introduce MAEL, a novel multi-agent collaboration framework that bridges this gap by integrating an experiential learning phase for continual cross-task experiential accumulation and a reward-guided experience retrieval mechanism for adaptive agent collaboration.
In MAEL, the agents reside on an undirected graph and exchange messages along the edges for task-solving workflow execution. The task-solving workflow combines a divide‑and‑conquer strategy with a solver‑critique collaboration pattern, guiding agents through task decomposition, subtask solving, and iterative solution refinement.
Concretely, during the experiential learning phase, agents collaboratively perform forward passes to decompose and solve tasks while quantifying the quality of each action as a reward. These agent-specific experiences, comprising input–output pairs annotated with rewards, are then stored and refined through a backward pass phase to augment the corresponding agents' distributed experience pools. At inference, each agent retrieves high-quality, task-relevant experiences via a reward-weighted similarity retrieval mechanism at each step, which are utilized as few shot examples to guide both the decomposition strategy and inter-agent communication.

We evaluate MAEL on a range of datasets spanning arithmetic reasoning (GSM \citet{Cobbe2021Training}), multi-domain question answering (MMLU \citet{Hendrycks2020MeasuringMM}), module-level code synthesis (HumanEval \citet{chen2021evaluating}), long-form generation (CommonGen-Hard \citet{Madaan2023Self}), and project-level code generation (SRDD \citet{qian2023communicative}).  
Our experiments demonstrate that MAEL consistently outperforms state‑of‑the‑art single‑agent and multi‑agent baselines.
By retrieving task-relevant high‑quality experiences at each reasoning step, MAEL guides agents away from erroneous or low‑quality solution paths, thereby reducing wasted exploration and accelerating convergence to high‑quality solutions.

In summary, our key contributions include:
\begin{itemize}[left=0pt, noitemsep]
\item We introduce a cross-task experiential learning paradigm for multi-agent cooperation, enabling agents to learn from past task experiences to improve the efficiency and quality of future task solutions.
\item We design MAEL, an MAS framework that integrates the divide-and-conquer paradigm with a solver-critique collaboration pattern, and incorporates experience learning and retrieval to guide task execution. This unified approach enables agents to decompose tasks, iteratively refine solutions through critique, and leverage past experiences for improved performance and convergence speed. 
\item We conduct extensive experiments to validate the effectiveness and efficiency of our proposed approach. Specifically, our experiments demonstrate that experience-driven retrieval significantly elevates task performance and convergence speed, and that MAEL generalizes well across diverse domains.
\end{itemize}

\section{Method}
\label{sec:method}
\begin{figure}
  \centering
  \includegraphics[width=1.0\textwidth]{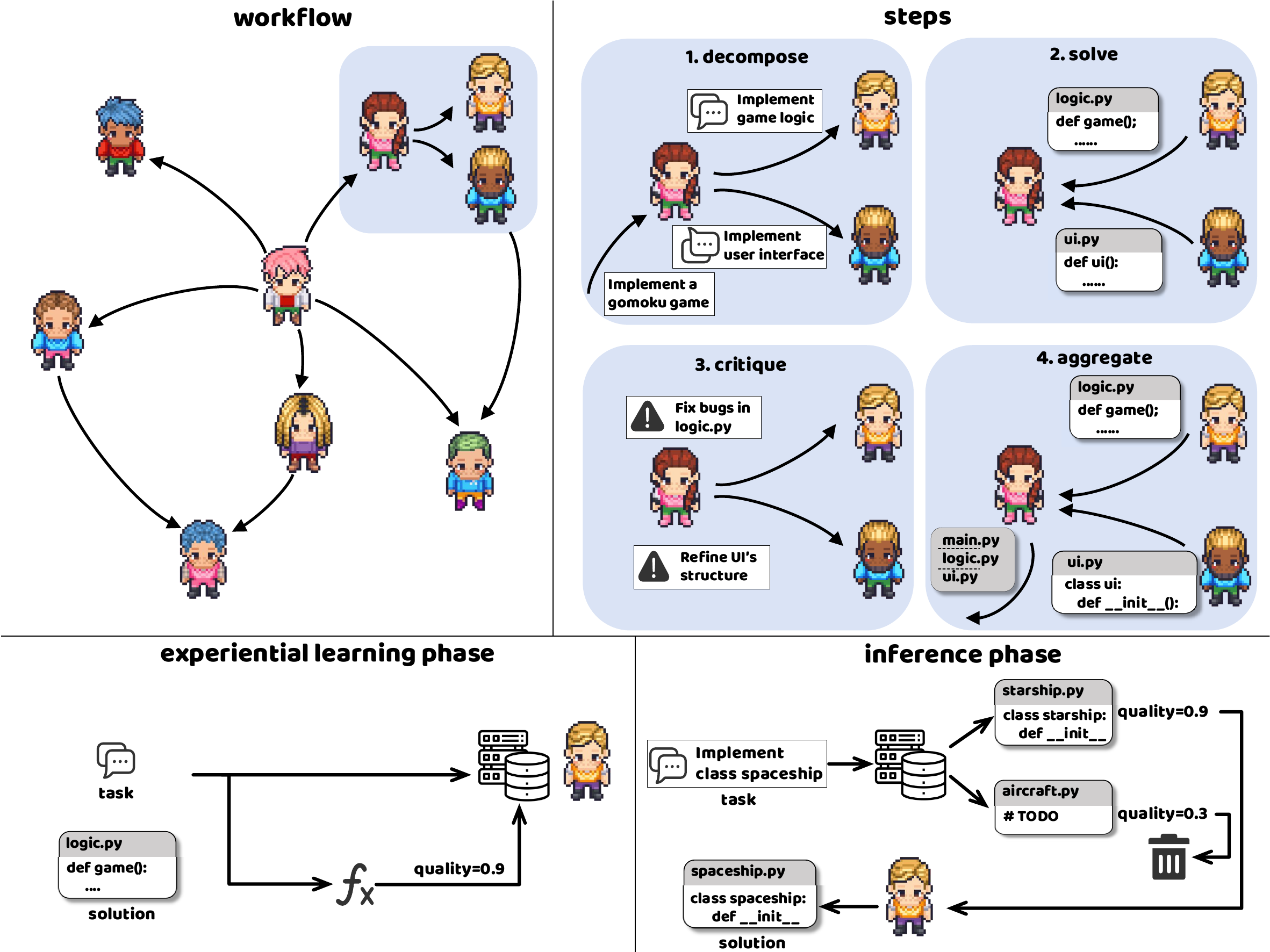}
  \caption{MAEL casts MAS as a graph of LLM-based agents, each endowed with an experience pool acting as learnable weights. During the experiential learning phase, agents execute tasks via a divide-and-conquer plus critique forward pass, collect state–action pairs annotated with rewards at every key decision step, and then perform a backpropagation-style update by storing quantified experiences into their own experience pools. In the inference phase, incoming tasks are solved through the same task-solving workflow, but each agent augments its reasoning with retrieved experiences—selecting high-quality task-relevant past experiences based on a combined similarity-and-reward score—as few-shot examples, guiding subtask decomposition, solution, refinement, and aggregation.}
  \label{fig:overview}
\end{figure}
In this section, we describe the architecture and learning procedures of MAEL, our cross‐task multi‐agent cooperation framework. As illustrated in Figure~\ref{fig:overview} , we view an MAS as a graph of LLM‐based agents that jointly decompose, solve, and refine subtasks. MAEL consists of two main stages: (1) an \emph{Experiential Learning Phase}, in which we collect and quantify agent interaction traces as experiences across a training corpus; and (2) an \emph{Inference Phase}, where we retrieve and integrate the most relevant high-quality past experiences to guide agent reasoning on new tasks.

\subsection{Multi-Agent Reasoning Workflow}
\label{sec:formulation}
MAS is a computational graph structure, with LLM-based agents serving as interconnected nodes that coordinate to accomplish complex tasks.
Formally, MAS can be denoted as an undirected graph $\mathcal{A} = (\mathcal{V}, \mathcal{E})$, where $\mathcal{V} = \{v_i\}^{N}_{i=1}$ is the set of $N$ nodes, and $\mathcal{E} = \{e_{i,j}\}, i,j\in[1,...,N ]$ is the set of undirected edges.
The $i$-th node $v_i$ represents the $i$-th agent with its associated experience pool $p_i$.
The edge $e_{i,j}$ represents the interaction between $i$-th and $j$-th agent in the agentic workflow. 
Drawing an analogy to traditional neural networks, agents $\mathcal{V}$ function like neurons, where their experience pools $\mathcal{P}$ act as learnable weights, and the agentic workflow $f(\cdot)$ defines their connections.

Existing multi-agent systems typically organize agents in linear or graph structures, achieving collaboration between adjacent agents in a streamlined workflow \citet{qian2023communicative, qian2023experiential}. However, for complex multi-phase tasks like software development, each agent must independently handle all subtasks within its phase, demanding strong long-context comprehension and generation capabilities.
To overcome the challenge, building upon the multi-agent network described above, we propose a novel divide-and-conquer workflow that supports adaptive task decomposition based on either functional module or pipeline and optimizes solutions through inter-agent critique and refinement.
The task-solving procedure is outlined as follows:

Starting from the agent with the highest closeness centrality in the multi-agent network, we execute the following recursive procedure:
    \emph{(1) Decompose}: The agent first checks if it can solve the task alone. 
    If \emph{no}, it decomposes the task into subtasks and assigns them to neighboring nodes. 
    \emph{(2) Solve}: If \emph{yes}, the agent solves the task alone and returns it to its assigner. 
    \emph{(3) Critique}: Once sub-solutions arrive from neighboring agents, the agent verifies their adequacy, delivers critiques or refinement suggestions for any inadequate ones, and returns the subtasks—along with the feedback—to the respective neighbor. 
    This process repeats until all subtasks are solved or the maximum rounds are reached.
    \emph{(4) Aggregate}: Finally, the agent consolidates all sub-solutions to produce the final solution for the task and returns it to its assigner. 
    The formalized algorithm of reasoning workflow can be found in Algorithm~\ref{alg:inference}.

\subsection{Experiential Learning Phase}
\label{sec:training}
Analogous to neural network training, for each input $x_i$ from the training dataset $\mathcal{D}=\{(x_i, y_i)\}_{i=1}^{N}$, we obtain the final solution through the agentic workflow as $\hat{y}_i = f(x_i)$.  
During this process, we quantify the performance of each reasoning step in the workflow by assigning step-level rewards using a task-specific metric, denoted as $r_t \leftarrow g(s_t, a_t)$, where $s_t$ denotes the input to the agent at decision step $t$, $a_t$ is the agent's output, and $r_t$ is the corresponding reward signal. 
These rewards are used to construct agent-specific experiences of the form $(s_t, a_t, r_t)$. 
The resulting experiences are stored in the corresponding agent's experience pool $\mathcal{P} = {(s_t, a_t, r_t)}_{t=1}^{T}$, which collectively serve as the learnable parameter space $\theta$ for the multi-agent system.  
The experience pool is incrementally constructed over the entire training dataset $\mathcal{D}$. In the following two sections, we provide a more detailed description of the forward and backward passes.

\paragraph{Forward Pass for Experience Formation}
\label{sec:forwardpass}
Given a task, the multi-agent network performs a forward pass i.e. executing the agentic workflow to generate the final output.
During the process, each agent takes in subtask descriptions and outputs responses for those decision steps.
Distinct reward calculation methods are applied on different critical decision steps(\emph{task solve}, \emph{task solvability judge}, \emph{task decompose}, \emph{solution critique} and \emph{solution aggregate}) to quantify each step's quality:
(1) $R_\mathrm{solve}$ is measured by the quality of the solution $\mathcal{R}(solution)$, with task-specific metric detailed in \cref{sec:metric};
(2) $R_{\text{solvability\_judge}}$ is measured by the quality of the solution $\mathcal{R}$ if the task is solved alone. Otherwise, it is measured by the average reward of the decomposed subtasks i.e. $\frac{1}{n} \sum_{i=1}^n R_{\text{subtask}_i}$;
(3) $R_{\text{decompose}}$ is measured by the average reward of the decomposed subtasks i.e. $\frac{1}{n} \sum_{i=1}^n R_{\text{subtask}_i}$;
(4) $R_{\text{critique}}$ is measured by the quality improvement of the solution after refinement, with larger improvement denoting better refinement;
(5) $R_{\text{aggregation}}$ is measured by the quality of the aggregated solution.

\iffalse
\begin{equation}
\begin{aligned}
& R_{\text{solve}} = \mathcal{R}(solution) \\
& R_{\text{solvability\_judge}} = 
\begin{cases} 
\mathcal{R}(solution), & \text{if solved alone} \\
\frac{1}{n} \sum\limits_{i=1}^n \mathcal{R}(sub\_solution_i), & \text{if decomposed into } n \text{ subtasks}
\end{cases} \\
& R_{\text{decompose}} = \frac{1}{n} \sum\limits_{i=1}^n \mathcal{R}(sub\_solution_i) \\
& R_{\text{critique}} = \mathcal{R}(solution_{refined}) - \mathcal{R}(solution_{original}) \\
& R_{\text{aggregate}} = \mathcal{R}(solution_{aggregated}) \\[5pt]
\end{aligned}
\end{equation}
\fi
\iffalse
\begin{table}[h]
\centering
\caption{Reward Calculation for Different Decision Points.}
\begin{tabular}{|c|c|c|} 
\hline
Decision Point & Prompt & Reward Calculation \\
\hline
Solvability Check &  & \\
\hline
Task Decomposition &  &  \\
\hline
Subtask Solution &  &  \\
\hline
Subtask Refinement & \\
\hline
Aggregation & \\
\hline
\end{tabular}
\label{tab:example}
\end{table}
\fi

\paragraph{Backward Pass for Experience Pool Update}
After deriving the reward of each critical decision step, we treat the agent's input-output pairs with rewards as experiences and store them in the corresponding agent's experience pool:
\begin{equation}
\mathcal{P} \gets \mathcal{P} \cup {(s_t, a_t, r_t)}, \quad \text{for } t=1,...,T
\label{eq:experience_update}
\end{equation}

\subsection{Inference Phase}
\label{sec:inference}
During the inference phase, the multi-agent network tackles a given task through the workflow.
For an individual agent $v_i$ at decision step $t$ during the execution, it retrieves task-relevant and high-quality experiences from the experience pool constructed in the experiential learning phase to serve as few shot examples, assisting in the task solution process.
The retrieval-augmented generation works as follows:
(1) For a given agent input $s_t$, we first compute its embedding using the text-embedding-ada-002 model;
(2) We then calculate the cosine similarities between this input embedding and the embeddings of existing experience entries in the experience pool $\{(s_j, a_j, r_j)\}_{j=1}^{\mathcal{N}}$ where $\mathcal{N}$ is the size of the agent's experience pool;
(3) The final retrieval score is computed as a weighted combination of the similarity score and its corresponding reward:
\begin{equation}
\mathrm{retrieval\_score} = \alpha \cdot \mathrm{similarity}(s_t, s_j) + (1-\alpha) \cdot \mathrm{reward}(s_j) \label{eq:scoring}
\end{equation}
where $\alpha$ is the weight and is set as 0.5 if not specified;
(4) The top-ranked experience entry $(s_j, a_j, r_j)$ is retrieved and is appended to the agent input $s_t$; 
(5) The augmented input is then fed to the agent to generate the response $a_t$.

We explore three experience retrieval strategies that differ in granularity and reliance on past experience:
\textbf{ (1) No experience} (denoted as MAEL\textsubscript{ØExp}), which performs a reasoning workflow solely based on the parameters learned by the model without consulting any stored experience.
\textbf{(2) Task-wise retrieval} (denoted as MAEL\textsubscript{Task}), which retrieves a complete, high-quality solution trace from the experience pool using the original task description, and consistently applies experiences in this single historical trace to guide the entire task-solving process.
\textbf{(3) Step-wise retrieval} (denoted as MAEL\textsubscript{Step}), which retrieves different high-reward experiences for each decision step, ensuring that agents leverage the most relevant high-quality experience at every step.

\section{Experiments}
\label{sec:experiment}
\subsection{Experimental Setting}
\label{sec:metric}
\paragraph{Baselines} We compare our method to the following single‑agent and multi‑agent approaches:
\begin{itemize}[left=0pt, noitemsep]
    \item \textbf{GPT-4o-mini} \citet{openai2024gpt4omini} is a compact, cost‑effective variant of OpenAI’s GPT‑4 architecture, chosen as a strong single‑agent baseline that balances inference speed and generation quality.
    \item \textbf{Claude-3.5-sonnet} \citet{anthropic2024claude35sonnet} is Anthropic’s advanced AI model excelling in graduate‑level reasoning (GPQA), code generation (HumanEval), and vision tasks, selected to represent state‑of‑the‑art few‑shot reasoning performance.
    \item \textbf{AutoGen} \citet{wu2023autogen} is a flexible MAS framework supporting diverse conversation patterns, tool integration, and customizable termination criteria, included as a benchmark for dynamic agent coordination.
    \item \textbf{EvoMac} \citet{hu2025selfevolving} is a self‑evolving MAS framework for software development that dynamically adapts agent roles and workflows using environmental feedback, serving as a cutting‑edge multi-agent baseline for iteration‑driven code refinement .
\end{itemize}

\paragraph{Datasets and Metrics}
To comprehensively evaluate our multi-agent method across a range of reasoning and generation challenges, we select five publicly available benchmarks that cover general knowledge, arithmetic reasoning, code synthesis, software development, and commonsense generation. Each dataset targets a distinct capability, and together they form a holistic testbed for heterogeneous downstream scenarios:

\begin{itemize}[left=0pt, noitemsep]
    \item \textbf{MMLU} \citet{Hendrycks2020MeasuringMM}: As one of the most extensive multi‑domain evaluation suites, MMLU measures broad world knowledge and logical inference across 57 subjects and varying difficulty levels. We adopt \textit{accuracy} to assess agents’ ability to retrieve and apply factual and conceptual information.  
    \item \textbf{GSM-8K} \citet{Cobbe2021Training}: GSM‑8K presents linguistically diverse grade‑school math problems requiring 2–8 reasoning steps, making it an ideal benchmark for multi‑step arithmetic reasoning. We use \textit{accuracy} to capture correct solution generation.  
    \item \textbf{HumanEval} \citet{chen2021evaluating}: This benchmark evaluates function‑level code generation on standardized test suites. We adopt \textit{pass@k} to measure the functional correctness of generated code, reflecting agents’ programming proficiency.    
    \item \textbf{CommonGen-Hard} \citet{Madaan2023Self}: This dataset challenges models to generate coherent sentences given unordered concept sets, probing contextual understanding, commonsense reasoning, and creative composition. We employ the official composite metric (grammar, fluency, relevance, logical consistency, and concept coverage) to gauge generative competence.
    \item \textbf{SRDD} \citet{qian2023communicative}: Capturing repository‑scale software development tasks from real‑world platforms, SRDD spans requirement understanding, design, code generation, and testing. We follow prior work in using \textit{completeness}, \textit{executability}, and \textit{consistency}—with the overall score as their product—to evaluate end‑to‑end development quality.
\end{itemize}

By combining these five datasets, we ensure our evaluation covers (1) factual and logical reasoning (MMLU), (2) arithmetic chain‑of‑thought (GSM‑8K), (3) code correctness (HumanEval), (4) commonsense generation (CommonGen‑Hard), and (5) large‑scale software workflow (SRDD), thus providing a thorough assessment of the proposed multi-agent framework.

\paragraph{Implementation Details}
By default, we employ a topology consisting of four nodes, aligning with other multi-agent baselines.
Unless otherwise specified, GPT-3.5 with temperature 0.2 is employed for interactive reasoning due to its optimal balance of efficacy and efficiency. 
For relatively simple datasets such as MMLU, HumanEval, GSM, we set maximum refinement round to 2 and maximum task decomposition layer to 1.
For relatively complex datasets such as CommonGen-Hard and SRDD, we 
set maximum refinement round to 3 and maximum task decomposition layer to 2.
We employ 30 training tasks for each dataset in our experiments. 
All baselines are rerun under identical settings.

\subsection{Overall Performance}
As illustrated in Table~\ref{tab:main}, MAEL outperforms other baselines on most datasets, validating its effectiveness.
MAEL\textsubscript{ØExp} surpasses gpt-4o-mini by 4.89 on average.
MAEL\textsubscript{Task} further improves the performance by 4.02.
Experience enhances performance when tasks share structural similarities.
The HumanEval, CommonGen-Hard, and SRDD datasets all show significantly improved performance when experience is incorporated.
We speculate that this could be due to the high overlap between task solution traces in these datasets.
For example, in SRDD, the task "Develop an expense management system" and task "Develop a library lending system" contain similar subtasks like "Develop a user management module" and "Develop a UI module".
This enables cross-task experiential learning, where high-quality past experiences can guide the solution of the current task, improving both effectiveness and efficiency.
However, on MMLU and GSM, performance degrades slightly when using experience.
We argue that this could be due to either the low overlap between task solutions or relatively low task complexity.
Therefore, when experience is introduced, the positive effect of reference is outweighed by the negative side effects of distraction caused by the experience and the increased difficulty of processing longer inputs.

\begin{table*}[t]
\centering
\setlength{\tabcolsep}{3pt}
\caption{Performance (\%) across various datasets, including both single-agent(
  \includegraphics[height=0.35cm]{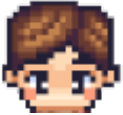}
) and multi-agent(
  \includegraphics[height=0.35cm]{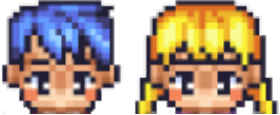}
) paradigms. For each dataset, the highest scores are highlighted in bold, while the second-highest scores are underlined.}
\label{tab:main}
{
  \definecolor{basecol}{RGB}{255,0,0}
  
  \newcommand{\colavg}[1]{
    \cellcolor{basecol!
      \fpeval{round((#1-50.08)/(75.00-50.08)*90 + 10, 0)}
    }#1
  }

  \begin{tabular}{l|c|cccccc}
    \toprule
    \textbf{Method}  
      & \textbf{Paradigm} & \textbf{MMLU} & \textbf{GSM}
      & \textbf{HumanEval} & \textbf{CommonGen}
      & \textbf{SRDD} & \textbf{Average} \\
    \midrule

    \rowcolor{gray!15}
    GPT-4o-mini  
      & \raisebox{-0.1\height}{\includegraphics[height=0.4cm]{sa.pdf}}
      & \textbf{83.33} & 23.33 & \textbf{90.00}
      & 37.05 & 70.19 & \colavg{60.78} \\

    Claude-3.5-sonnet  
      & \raisebox{-0.1\height}{\includegraphics[height=0.4cm]{sa.pdf}}
      & 30.00 & 53.33 & \textbf{90.00}
      & 37.73 & 55.72 & \colavg{53.36} \\

    \rowcolor{gray!15}
    AutoGen  
      & \raisebox{-0.1\height}{\includegraphics[height=0.4cm]{ma.pdf}}
      & \textbf{83.33} & 23.33 & 80.00
      & 33.03 & 56.21 & \colavg{55.18} \\

    EvoMac  
      & \raisebox{-0.1\height}{\includegraphics[height=0.4cm]{ma.pdf}}
      & \underline{80.00} & \textbf{70.00} & 46.67
      & \underline{45.48} & 8.26 & \colavg{50.08} \\

    \midrule

    \rowcolor{gray!15}
    MAEL\textsubscript{ØExp}  
      & \raisebox{-0.1\height}{\includegraphics[height=0.4cm]{ma.pdf}}
      & \underline{80.00} & \textbf{70.00} & 76.67
      & 34.50 & 67.17 & \colavg{65.67} \\

    MAEL\textsubscript{Task}  
      & \raisebox{-0.1\height}{\includegraphics[height=0.4cm]{ma.pdf}}
      & \underline{80.00} & \underline{60.00} & \textbf{90.00}
      & \textbf{46.54} & \underline{71.93} & \textbf{\colavg{69.69}} \\

    \rowcolor{gray!15}
    MAEL\textsubscript{Step}  
      & \raisebox{-0.1\height}{\includegraphics[height=0.4cm]{ma.pdf}}
      & 76.67 & 50.00 & \underline{86.67}
      & 42.81 & \textbf{76.43} & \underline{\colavg{66.52}} \\

    \bottomrule
  \end{tabular}

}

\end{table*}

\subsection{Effect of Experience Retrieval and Utilization}
In Table~\ref{tab:quality}, we compare the performance of baseline AutoGen with MAEL using three different retrieval strategies on SRDD dataset.
MAEL\textsubscript{Task} differs from MAEL\textsubscript{Step} in that, it retrieves experiences based on the initial task description and uses it throughout the entire task while MAEL\textsubscript{Step} retrieves experiences independently at each step.
Among them, MAEL\textsubscript{Step} achieves the best performance, with a 9.2\% improvement compared to MAEL\textsubscript{ØExp}, and a 20.21\% gain over the AutoGen baseline.
Such improvement likely stems from the accumulation of locally optimized decisions across all steps, ultimately boosting overall performance.
\begin{table}[h]
\centering
\caption{Performance on SRDD dataset in terms of quality (\%).}
\label{tab:quality}
{ 

  \definecolor{basecol}{RGB}{255,0,0}

  \newcommand{\colavg}[1]{
    \cellcolor{basecol!
      \fpeval{round((#1-56.0)/(82.00-56.0)*90 + 10, 0)}
    }#1
  }

  \begin{tabular}{l|cccc}
    \toprule
       \textbf{Method} 
      & \textbf{Completeness} 
      & \textbf{Executability} 
      & \textbf{Consistency} 
      & \textbf{Quality} \\
    \midrule

    \rowcolor{gray!15}
    AutoGen                 
      & 85.0  & 85.0        & 80.0  
      & \colavg{56.0} \\
    \midrule

    MAEL\textsubscript{ØExp}
      & 90.0  & \textbf{95.0}
      & 78.9  
      & \colavg{67.2} \\

    \rowcolor{gray!15}
    MAEL\textsubscript{Step}
      & 95.0  & \textbf{95.0}
      & \textbf{80.2}
      & \textbf{\colavg{76.4}} \\

    MAEL\textsubscript{Task}
      & \textbf{100.0}
      & 90.0
      & 79.4
      & \colavg{71.6} \\
    \bottomrule
  \end{tabular}
} 

\end{table}

\begin{table}[t]
\centering
\caption{Performance on SRDD dataset in terms of scale and complexity.} 
\label{tab:scale}
{ 
  \begin{tabular}{l|ccccc}
    \toprule
       \textbf{Method} 
      & \textbf{Complexity($\uparrow$)} 
      & \textbf{\# Line($\uparrow$)} 
      & \textbf{\# Token($\uparrow$)} 
      & \textbf{\# Function($\uparrow$)} 
      & \textbf{\# File($\uparrow$)} \\
    \midrule

    \rowcolor{gray!15}
    AutoGen                    
      & 11.20 &  55.70 & 503.40 &  5.10 & 1.00 \\
    \midrule

    MAEL\textsubscript{ØExp}  
      & 28.70 & 114.03 & 781.95 & 13.15 & 4.15 \\
    \rowcolor{gray!15}
    MAEL\textsubscript{Step}  
      & 27.55 & 119.30 & 805.65 & 14.05 & 4.20 \\
    MAEL\textsubscript{Task}  
      & 22.90 &  99.75 & 721.75 & 11.15 & 3.80 \\
    \bottomrule
  \end{tabular}
}

\end{table}

To gain a deeper understanding of the generated software projects, in addition to the above quality metric, we also calculate several metrics reflecting the overall scale and complexity of each software project. Specifically, for each project, we compute its total cyclomatic complexity, total number of non-commented lines of code, total token count, total number of functions, and total number of files. We then report the average values of these metrics across all projects. Higher values of these metrics indicate greater code scale and complexity.
As demonstrated in Table~\ref{tab:scale}, the software projects generated by MAEL\textsubscript{Step} not only possess higher quality but also exhibit higher complexity than others.
Our manual inspection reveals that introducing diverse references at each step enhances variability, resulting in more modular code.

\subsection{Efficiency Analysis}
\label{sec:efficiency}

\begin{figure}[h]
    \centering
    \includegraphics[width=1\linewidth]{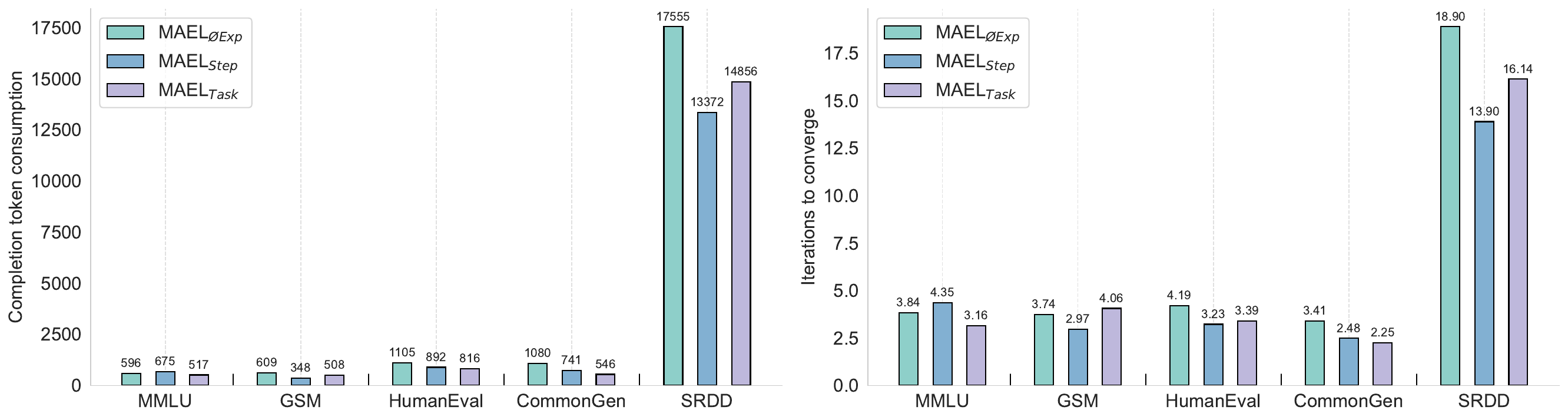}
    \caption{Comparison of token consumption (Left) and number of convergence round (Right). Since the SRDD dataset requires significantly more rounds and tokens than other datasets, for better visualization, we normalize all values relative to MAEL\textsubscript{ØExp}.}
    \label{fig:token_iteration}
\end{figure}

As shown in Figure~\ref{fig:token_iteration}, leveraging stored experience consistently yields significant reductions in both completion token consumption during inference and the number of iterations required for solution convergence across nearly all datasets.

\textbf{Token efficiency.} By guiding agents with high‑quality few‑shot exemplars, both MAEL\textsubscript{Task} and MAEL\textsubscript{Step} substantially reduce the need for repeated critique refinements. For instance, on the CommonGen‑Hard dataset, MAEL\textsubscript{Task} cuts token usage by 49\% compared to the MAEL\textsubscript{ØExp} baseline, while simultaneously boosting the overall quality score by 12.04 points—setting a new state of the art across all baselines.

\textbf{Convergence speed.} Introducing experience also lowers the likelihood of generating low‑quality or erroneous outputs, thereby reducing the number of dialogue turns required to refine solutions. On the SRDD dataset, MAEL\textsubscript{Step} reduces the average dialogue rounds to convergence by 26\%, and yields a 9.25‑point gain in software quality—again outperforming every baseline.

These findings demonstrate that experience-augmented reasoning facilitates faster convergence of multi-agent workflows while simultaneously enhancing final solution quality, particularly on tasks with overlapping solution traces where cross‑task experience learning is most beneficial.

\subsection{Analysis of Factors Affecting Experience Retrieval}

This section investigates how the two key components of our retrieval score—reward and similarity—impact overall performance. We define four retrieval settings by modifying the scoring function
$\text{score} \;=\; \alpha \times Sim \;+\;(1-\alpha)\times R$, where \(R=reward \) for “High Reward” and \(R=1-reward\) for “Low Reward”, and \(Sim=similarity \) for “High Similarity” and \(Sim=1-similarity \) for “Low Similarity”.
\looseness=-1

Improving either reward or similarity yields substantial performance uplift, while combining both factors produces the largest gains. As shown in Table~\ref{tab:factor}, on HumanEval, using High Similarity alone increases performance from 0.800 to 0.867 (an 8.3\% relative gain), while using High Reward alone raises it to 0.867 (also 8.3\%). When both factors are optimized simultaneously (High Reward \& High Similarity), we observe a further boost to 0.90, corresponding to a 12.5\% gain over the baseline.

Retrieving experiences—regardless of quality—consistently outperforms no retrieval. Even the Low Reward \& Low Similarity condition—our most conservative retrieval strategy—outperforms the no‑experience baseline by 6.7\% on HumanEval, confirming the robustness of MAEL to noisy or weakly relevant experiences. A similar pattern emerges on CommonGen, where joint optimization yields a 10.1\% improvement and the Low Reward \& Low Similarity setting still achieves a 0.9\% absolute gain over no‑experience.

\begin{table}[ht]
  \centering
  \caption{Analysis of factors (\textit{i.e. quality and similarity}) affecting retrieval. Performance (\%) of MAEL\textsubscript{ØExp} is 73.3 and 34.5 on HumanEval and CommonGen respectively.}
{ 
  \rowcolors{2}{gray!15}{white}

  \begin{tabular}{lcc|cc}
    \toprule
      & \multicolumn{2}{c}{HumanEval} 
      & \multicolumn{2}{c}{CommonGen} \\
    \cmidrule(lr){2-3} \cmidrule(lr){4-5}
      & Low Reward & High Reward 
      & Low Reward & High Reward \\
    \midrule
    Low Similarity  
      & 80.0 & 86.7 
      & 35.4 & 36.2 \\
    High Similarity 
      & 86.7 & \textbf{90.0} 
      & 36.4 & \textbf{46.5} \\
    \bottomrule
  \end{tabular}
} 

  \label{tab:factor}
\end{table}

\subsection{Scaling Trend of Experience Pool}
In the preceding experiments, we demonstrated that the experience mechanism not only accelerates convergence toward high-quality solutions but also improves the overall quality of the generated solutions. In this section, we further investigate how performance scales with the size of the experience pool. 
\looseness=-1

\begin{figure}[h]
    \centering
    \includegraphics[width=0.6\linewidth]{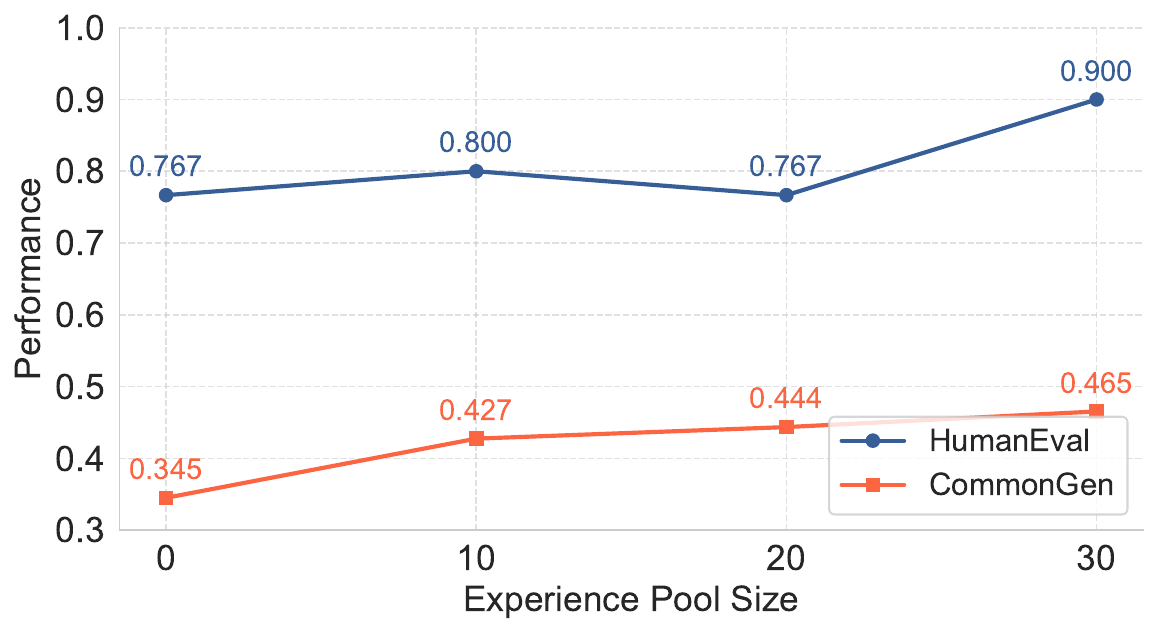}
    \caption{Performance of MAEL\textsubscript{Task} on HumanEval and CommonGen benchmarks with varying experience pool sizes (0, 10, 20, and 30 accumulated tasks).}
    \label{fig:scaling}
\end{figure}

As shown in Figure~\ref{fig:scaling}, on HumanEval, we observe a clear upward trend. Starting from a baseline of 76.7 \% when no prior experience is available, performance increases to 80.0 \% with 10 stored tasks, to 76.7 \% at 20 tasks (indicating a minor fluctuation), and ultimately reaches 90.0 \% at 30 tasks. The non-monotonic dip at 20 tasks suggests that the relevance or diversity of stored experiences, and not just their quantity, can influence retrieval quality.

On CommonGen, initial gains are more gradual: performance climbs from 34.5 \% with no experience to 42.7 \% at 10 tasks (+8.2 pp), 43.4 \% at 20 tasks (+0.7 pp), and finally 46.5 \% at 30 tasks (+3.1 pp). The gentler slope here reflects CommonGen’s larger solution space and higher variability; nevertheless, the overall 12.0 pp improvement highlights MAEL’s capacity to leverage prior collaborative episodes even in complex language generation settings, with the 30-task experience pool configuration surpassing all existing baselines and achieving new state-of-the-art performance on this benchmark.
\looseness=-1

\begin{figure}
  \centering
  \includegraphics[width=1\textwidth]{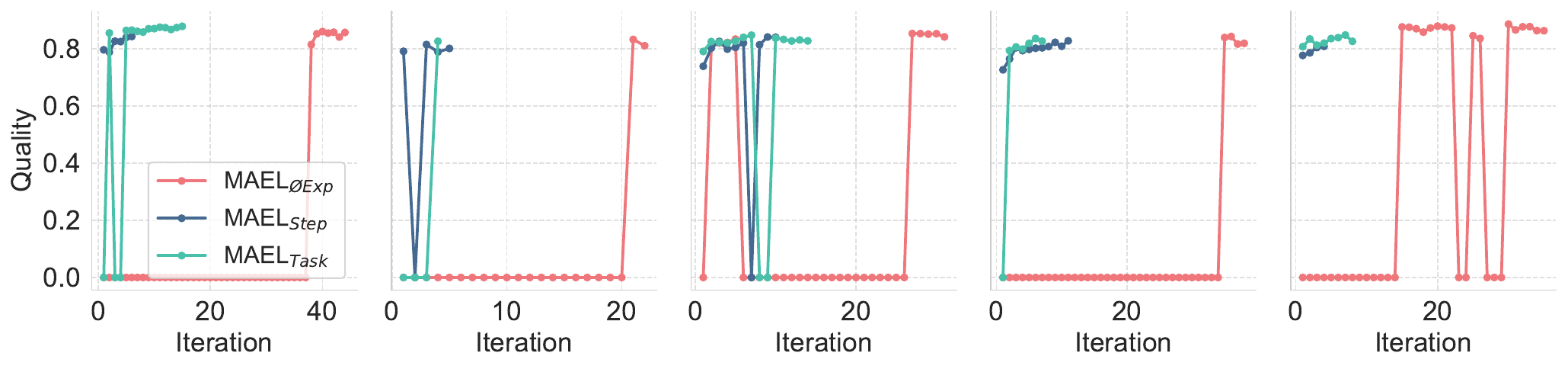}
  \caption{Solution quality comparison across five cases using different experience retrieval methods.}
  \label{fig:progression}
\end{figure}

\subsection{Case Study}
\label{sec:casestudy}
\begin{figure}[h]
  \centering
  \includegraphics[width=\linewidth]{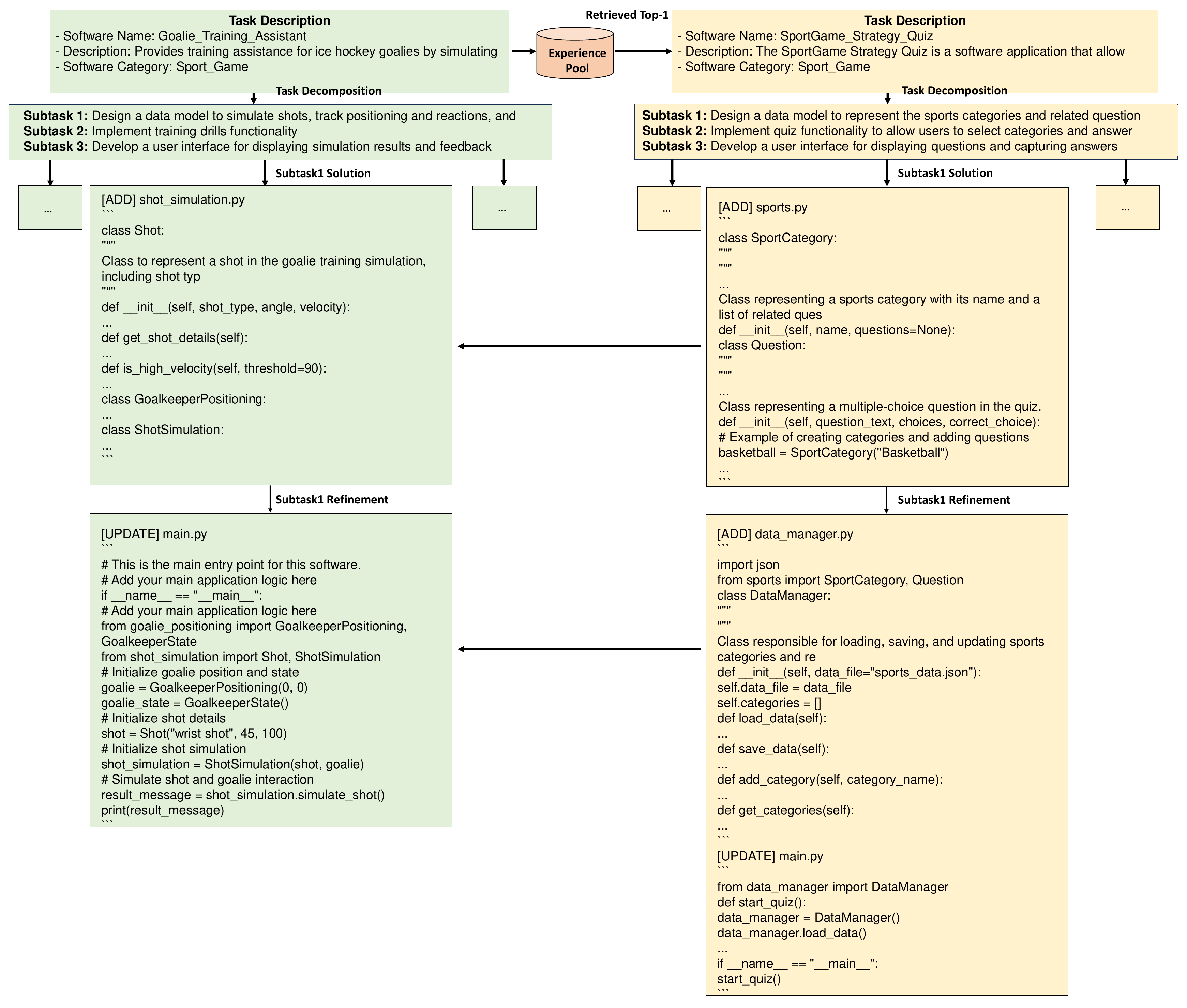}
  \caption{A case illustrating the assistance of experience (yellow).}
  \label{fig:case}
\end{figure}

Figure~\ref{fig:progression} illustrates the quality progression over iterations across five SRDD cases. 
As observed, the convergence steps required by MAEL\textsubscript{Task} and MAEL\textsubscript{Step} are significantly fewer than those of MAEL\textsubscript{ØExp}.
Figure~\ref{fig:case} demonstrates how the retrieved experiences assist the task-solving process using fourth case. 
In this case, it can be observed that MAEL\textsubscript{Task} retrieves the historical task from the experience pool based on similarity and quality.
Under the guidance of this experience, the task is decomposed into three subtasks. While addressing the first subtask, the agent, guided by experience, writes code for Shot, GoalkeeperPositioning, and ShotSimulation, but these have not yet been called in the main function, which still contains pass. At this point, the completeness is 0.
It can be seen that in the experience, the agent modified the main function by incorporating the newly implemented data\_manager class, resolving previously unimplemented parts. Using this experience as guidance, the agent responsible for the current subtask also modifies the main function, integrating the class implemented in the previous round into the main function logic and replacing the pass section. This results in the first version of the code with a completeness of 1.
At this stage, the code repository's completeness and executability both reach 1, while consistency is 0.79786 and quality is 0.79786.
Throughout this process, by leveraging experience retrieval, we resolved a code flaw in a single round of dialogue that would otherwise have required multiple rounds of critique without prior experience. This significantly improves the efficiency and performance of code generation.

\section{Related Work}
Large language models (LLMs), pretrained on massive text corpora to optimize billions of parameters, have become a cornerstone of modern natural language processing research \citep{Brown2024Large, bubeck2023sparks, vaswani2017attention, radford2019language, touvron2023llama, wei2022emergent, Shanahan2023, chen2021evaluating, brants2007large, ouyang2022training, yang2023large, qin2023large, kaplan2020scaling}. Building directly on these foundational advances, there has been a surge of interest in autonomous agents that harness the reasoning and generation strengths of LLMs \citep{zhou2023webarena, wang2023recagent, park2023generative, Wang2023Plan, AutoGPT, Wang2024Learning, shinn2024reflexion}. Such agents have demonstrated sophisticated memory mechanisms \citep{park2023generative, sumers2023cognitive}, dynamic planning abilities \citep{Liu2023Dynamic}, and effective tool-use strategies \citep{schick2023toolformer, cai2023large, qin2023toolllm, Ruan2024Observational, yang2023large}, empowering them to operate autonomously across complex, real-world tasks \citep{zhao2024expel, ma2023laser, zhang2023generative, Wang2023Augmenting, ding2023designgpt, weng2023llm}.
Concurrently, the autonomous communication among multiple agents have become a promising paradigm, signaling a shift towards MAS\citet{chen2023agentverse, Chen2024Internet, Cohen2023LM, Guo2024Embodied, li2023camel, wu2023autogen}.
Recently, researchers have begun exploring self-evolution in MAS \citet{hu2025selfevolving, zhang2025evoflow, yuan2024evoagent, shang2024agentsquare, liu2024dynamic, hu2024automated, feng2025heterogeneous}, as it enables agents to adapt to new tasks without extensive manual redesign.
EvoMAC \citep{hu2025selfevolving} iteratively
adapts agents and their connections during the test phase of each task via textual backpropagation for software development.
Our work differs from prior approaches by incorporating a training phase that accumulates historical task experience to facilitate knowledge transfer across tasks, while also assigning reward to each experience to enable finer-grained evolutionary optimization.

\section{Conclusion}
\label{sec:conclusion}
In this work, we have presented MAEL, a cross-task multi-agent collaboration framework that enriches traditional MAS pipelines with an explicit experience-accumulation learning stage and a reward-weighted experience retrieval mechanism. During experiential learning, agents perform divide-and-conquer plus critique forward passes, obtain fine-grained reward annotations for each action, and update individualized experience pools via backward passes; at inference, they retrieve high-quality, task-relevant experiences to guide each task-solving step. MAEL consistently outperforms state-of-the-art single- and multi-agent methods across various datasets. Ablations show that global experience retrieval with iterative critique boosts solution quality, speeds convergence, and enriches code complexity, while the divide-and-conquer plus critique workflow balances computational load and curbs hallucination. We hope this work will inspire more efficient multi-agent collaboration and improved cross-task generalization in future research.

\bibliographystyle{unsrtnat}
\bibliography{neurips_2025}

\end{document}